\pdfoutput=1

\documentclass[10pt, a4paper, dvipsnames]{article}

\usepackage{lrec-coling2024} %

\usepackage{booktabs}
\usepackage{textcomp} %
\usepackage{fixltx2e}
\usepackage{makecell}
\usepackage{longtable}

\title{QueryNER: Segmentation of E-commerce Queries*\thanks{* This is a preprint version of this article. Please cite the version published in the \href{https://aclanthology.org/}{LREC-COLING 2024 proceedings in the ACL Anthology}.}}

\name{Chester Palen-Michel$^{\ast}$, Lizzie Liang$^{\dagger}$, Zhe Wu$^{\dagger}$, Constantine Lignos$^{\ast}$} 

\address{$^{\ast}$Michtom School of Computer Science, Brandeis University\\
    $^{\dagger}$eBay Inc.\\
     \texttt{\{cpalenmichel,lignos\}@brandeis.edu}\\
     \texttt{\{lliang,zwu1\}@ebay.com}\\
}

\abstract{
We present QueryNER, a manually-annotated dataset and accompanying model for e-commerce query segmentation. 
Prior work in sequence labeling for e-commerce has largely addressed aspect-value extraction which focuses on extracting portions of a product title or query for narrowly defined aspects. 
Our work instead focuses on the goal of dividing a query into meaningful chunks with broadly applicable types. 
We report baseline tagging results and conduct experiments comparing token and entity dropping for null and low recall query recovery. 
Challenging test sets are created using automatic transformations and show how simple data augmentation techniques can make the models more robust to noise. 
We make the QueryNER dataset publicly available. 
 \\ \newline }

\begin{document}

\maketitleabstract

\section{Introduction}

An important challenge in e-commerce query understanding is returning relevant results for null and low recall queries. 
These queries return few or no results due to vocabulary mismatch or queries containing too many terms that over-constrain the retrieval process.
A common approach to recover from null and low recall queries is to rewrite terms with similar words or to remove terms from the query to relax the constraints.  
In applying these recovery methods, queries are often treated as unstructured sequences of tokens \citep{bilotti2004query,chen2009query,tan2017query,gamzu2020query} rather than natural groupings of tokens.
Past work in natural language processing (NLP) has treated grouping tokens as a shallow parsing or chunking task \citep{abney1992parsing,ramshaw-marcus-1995-text,molina2002shallow}.
By chunking a query, we can find the boundaries between spans of tokens and identify the purpose of each span. 
This allows us to weight spans rather than tokens, drop spans as a recovery approach, and potentially better cluster similar chunks and link them to a knowledge graph. 

Chunking is often framed as a sequence labeling task, and while there has been sequence labeling work in e-commerce, it has largely focused on aspect-value extraction. 
Aspect-value extraction identifies portions of a string of text (values) for more narrowly defined aspects like brand or color (\citealt{joshi-etal-2015-distributed};\citealtlanguageresource{papenmeier2021dataset}).
While aspect-value extraction does identify some natural groupings of tokens, the goal is often only to identify spans that are values for predefined aspects. 
If there is no aspect defined for a span of text, it will not be identified. 
Aspect-value extraction approaches tend to have either few aspect types with many tokens not included as part of a span, or they have large complex aspect ontologies  with thousands of aspects \citep{xu-etal-2019-scaling}. 
While there is work on e-commerce aspect-value extraction, there have been few datasets released publicly for research.
Much of the broader sequence labeling datasets that are publicly available to NLP researchers focus on the task of Named Entity Recognition (NER). 
NER datasets typically include general entity types like persons, organizations, and locations (e.g. \citealtlanguageresource{tjong-kim-sang-2002-introduction,sang2003introduction,hovy-etal-2006-ontonotes}) but not the entity types associated with spans for e-commerce. 
E-commerce data presents additional challenges compared to other sequence labeling datasets since it can be more noisy and unstructured \citep{putthividhya-hu-2011-bootstrapped,zirikly-hagiwara-2015-cross}.

\begin{table*}[tb]
\begin{center}
\resizebox{\linewidth}{!}{
\begin{tabular}{c}
\toprule
Aspect-Value Extraction:\\
\texttt{High - end} \texttt{\textcolor{blue}{[speaker cover]}} \texttt{for}   \texttt{\textcolor{CadetBlue}{[B \& W]}}   \texttt{\textcolor{RedOrange}{[805d]}}  \texttt{1 pair}  \texttt{made of \textcolor{Fuchsia}{[velvet]}  \textcolor{OliveGreen}{[suede]}} \texttt{made to order} \\
\midrule
QueryNER Segmentation:\\
\texttt{\textcolor{TealBlue}{[High - end]}}  \texttt{\textcolor{blue}{[speaker cover]}}   \texttt{for}   \texttt{\textcolor{CadetBlue}{[B \& W]}   \textcolor{RedOrange}{[805d]}  \textcolor{RedViolet}{[1 pair]}  \textcolor{Fuchsia}{[made of velvet]}   \textcolor{OliveGreen}{[suede]}}  \texttt{\textcolor{Brown}{[made to order]}} \\

\bottomrule
\end{tabular}
}
    \caption{
    An example of a query segmented with a hypothetical aspect-value extraction ontology compared with QueryNER's chunking ontology. 
    }
    \label{tab:query-example}
    \end{center}
\end{table*}

We present QueryNER, a publicly available dataset,\footnote{\url{https://github.com/bltlab/query-ner}} manually annotated for e-commerce query segmentation. 
The task in QueryNER is not to extract aspects, but rather to segment the user’s query into meaningful chunks. 
Unlike the e-commerce task of aspect extraction, which tends to focus on fine-grained types that are often specific to particular categories of items, the tag types of QueryNER aim to be broadly applicable to queries for any product category. 
This difference in approach leads QueryNER to have nearly all tokens included in some form of span, with the exception of a few special characters, some prepositions and conjunctions.
In Table~\ref{tab:query-example}, an example query is shown with the entity spans identified following the QueryNER schema compared with a hypothetical aspect-value extraction. 
The type ontology is intended to be a small number of entity types and general purpose enough that it can be used for a broad range of e-commerce product categories. 
Annotators also do not necessarily need to become domain experts in the products involved in the annotation process or familiarize themselves with thousands of aspects.

As seen in Table \ref{tab:query-example}, \textit{Speaker cover}, \textit{made of velvet}, \textit{made to order}, \textit{1 pair} and \textit{high-end} are all natural chunks of the query. 
For example, \textit{made of velvet} clearly refers to the material while \textit{made to order} is a more general description of the product and may very well not be covered under certain aspect ontologies.

\textbf{Contributions:  }
Our contributions are the following. 
(1) We define a type ontology and annotation guidelines that are broadly applicable to e-commerce segmentation. 
(2) We release QueryNER, a new manually annotated dataset and open benchmark for this task.
(3) We report baseline results from models trained on the QueryNER dataset.
(4) We discuss the results of an experiment showing promising directions for using QueryNER as part of a null and low query recovery strategy by dropping spans rather than individual tokens.  
(5) We conduct experiments showing benefit of data augmentation for query segmentation.

\section{Related Work}

Sequence labeling is a well established task in NLP with tasks like NER (e.g. \citealtlanguageresource{tjong-kim-sang-2002-introduction,sang2003introduction,hovy-etal-2006-ontonotes}) and chunking (e.g. \citealt{abney1992parsing,ramshaw-marcus-1995-text,molina2002shallow}) framed as labeling each token with a label indicating whether it is part of a span or not and what type of span. 
The labeling most typically uses BIO labels where B marks the beginning of a span, I marks inside a span and O, outside the span. 
Other label encodings have been used such as BIOES \citep{radford-etal-2015-named}. 

Prior work in aspect-value extraction has largely framed the task as sequence tagging as well.
\citet{joshi-etal-2015-distributed} experimented with embedding representations for aspect-value extraction. 
\citet{farzana2023knowledge} made use of a knowledge graph and entity linking to reformulate queries and use aspect-value extraction for entity span information in a rephrasing model. 
However, neither of these works release a public dataset. 

Since there is a lack of public aspect-value extraction datasets, there has been work with alternative approaches to create training datasets such as distant supervision or iterative bootstrapping approaches.  
\citet{zhang-etal-2020-bootstrapping} examine a bootstrapping method using positive unlabeled learning. They point out that while there are NER datasets for PER, ORG, LOC, there are not many publicly available NER datasets for e-commerce related tasks.
\citet{putthividhya-hu-2011-bootstrapped} work with product titles and bootstrap from a seed list of product attributes. 
\citet{xu-etal-2019-scaling} use distant supervision to automatically create training data for a limited number of  categories. 
They use a question-answering approach for aspect-value extraction. 

Some e-commerce sequence labeling data has been released, but it has some drawbacks. 
\citetlanguageresource{papenmeier2021dataset} release an e-commerce dataset for attribute-value extraction, but it only covers queries about laptops and jackets.
Due to the nature of e-commerce work, even datasets with general purpose queries or product titles are often not publicly available. 
\citetlanguageresource{reddy2023shopping}
 created a dataset of Amazon e-commerce queries and matching product titles including judgments for relevance using ESCI labels (exact match, substitute, complement, irrelevant).

\section{Dataset Creation}
QueryNER uses a subset of the Amazon Shopping Queries Dataset \citeplanguageresource{reddy2023shopping} as the underlying data.
We release our dataset as token offsets that can be mapped to the Shopping Queries Dataset.
QueryNER consists of an ontology of 
17 types.
One main difference in the guidelines given to annotators was to mark the fullest extent of a span possible. 
This intended to include words like \textit{size} in the span \textit{[size 12]} rather than \textit{size [12]}.
QueryNER follows the CoNLL tradition of using BIO format.

\subsection{Ontology and Annotation Guidelines}

The following are descriptions and examples for each entity type in QueryNER:

\textbf{core\_product\_type:} 
The main thing being sold. 
Generic ways of describing a product. 
These are not official product names but common objects. 
Examples: teapot, tennis shoes, figurine, lounge pants, dish soap

\textbf{product\_name:} 
The specific name of a product or model name. 
Examples: F150, air jordan 7, sorento

\textbf{product\_number:} 
The number for a product. 
It can be e-commerce product number or companies product number. 
Editions of an item that are numbered can also be marked as product number as well as trading card or comic numbers. 
Examples: BQ4422-001, 7101, DCC-3200P1

\textbf{modifier:}
Modifier is used for spans that clarify the type of product. 
This can describe certain features a project has like ``2 in 1" or ``high performance".
Modifier can also be used for constraining the type of a product.
Modifier can also be used as a catch all for ``type" of a product that does not fit in other predefined categories. 
For example, ``for sale" is not quite a condition nor a price. 
Similarly ``fast shipping", ``trusted seller" may not fit other categories, but are still meaningful chunks. 

\textbf{creator:}
The company or person who creates or produces the product. 
It could also be the designer name associated with the product or brand name. 
Examples:ford, disney, jim shore, honda, Hot wheels, dc comics

\textbf{condition:} 
The condition of the product. 
This describes whether the product is new or old and can go into more detail about things such as whether a product includes its original tags. 
Examples: new, used, mint condition

\textbf{UoM (Unit of Measurement):}
Any way of measuring size or other unit of measurement. 
This can include everything from clothing sizes, to lengths and widths, car engine sizes, battery capacity, amount of memory in a computer, lens sizes for cameras. 
This includes time expressions that are units of measurement such as 30 minutes or 4 hours of battery life.

\textbf{department:} 
Category of the population the item was made for. 
Examples: Mens, womens, kids, jr., wmns

\textbf{material:} 
The material or physical entity that makes up the item. 
Examples: denim, canvas, plastic, metal, cotton, felt

\textbf{time:}
An expression of the date or time associated with the product that is not a unit of measurement. 
For example, 30 mins or 4 hours for battery life should be labeled UoM.
Time spans such as 1920-1924 are marked as a single span [1920-1924].

\textbf{content:} 
Names of characters, titles of tv or movies, sayings or phrases that appear on or within the product itself.
Many mugs, t-shirts, figurines, or comic books have some form of content or characters associated with them.

\textbf{color:} 
The color, pattern, appearance related to the surface appearance or 2-dimensional design of the product. 
Examples: unc blue, light gray, wolf gray, floral, cherry

\textbf{shape:} 
The shape, form, or positioning or 3-dimensional design of the product. 
Includes design descriptions for things like clothing, accessories, or automotive that refer to 3-dimensional descriptions of the item.
Examples: fit, slim, low, long sleeve, flat, rear, front, rectangular, orb

\textbf{quantity:}
The number of the product being sold. Includes for example ``lot of 4".
Examples: 2 cds, lot of 4, multi-lot, package of 6, 3-box break

\textbf{occasion:}
The purpose or intended use of an item. 
Typically an event, holiday, season, or occasion. ``hiking boots" is its own product and `hiking` in this case should NOT be marked as a occasion.
Examples: sport, athletic, wedding, winter, halloween, bridal, birthday

\textbf{origin:} 
The origin of a product. 
This is likely the location it comes from but could also be a specific event where the item was created such as a convention. 
States or provinces can also be an origin tag.

\textbf{price:} 
The price of a product. 
Also includes words expressing relative price like ``expensive", ``cheap", ``lowest price", or ``good deal".

A full copy of the annotation guidelines is included in Appendix~\ref{sec:guidelines}.

\subsection{Annotation Process}
We began with pre-liminary annotation experiments on internal data in multiple categories in order to refine the annotation guidelines and type ontology. 
We then turned to conducting annotation on a subsection of the public Shopping Queries Dataset.
Three annotators were assigned to the test data in order to assess agreement and ensure quality. 
One annotator was assigned to the training and development portions of the data.
Additional quality checks were conducted which included flagging queries without a core product type or multiple core product types for further review. 
We originally selected ten thousand queries for annotation. 
Some were thrown out due to being outside the target language. 
Some were removed for profanity after not being identified in the original filtering and query selection.
The test set was further adjudicated to resolve conflicts and review annotation. 
The adjudication process generally accepted annotations where more than one annotator was in agreement unless it appeared to be a clear violation of the annotation guidelines. 
The adjudication was conducted by a single adjudicator who was involved in the creation of the annotation guidelines.

\subsection{Agreement}
\begin{table*}[tb]
\small
\centering
\begin{tabular}{@{}lrrrrrrrr@{}}
\toprule
 & \multicolumn{4}{c}{\textbf{Entity Level Agreement}} & \multicolumn{4}{c}{\textbf{Token Level Agreement}} \\ 
 \cmidrule(lr){2-5} \cmidrule(l){6-9} 
\textbf{Dataset} & Fleiss & Cohen 1-2 & Cohen 1-3 & Cohen 2-3 & Fleiss & Cohen 1-2 & Cohen 1-3 & Cohen 2-3 \\
\cmidrule(r){0-0}  \cmidrule(lr){2-5} \cmidrule(l){6-9} 
Internal trial & 65.6 & 59.6 & 78.1 & 59.0 & 77.2 & 73.4 & 84.6 & 73.7 \\
QueryNER & 38.4 & 38.7 & 37.2 & 39.4 & 59.4 & 56.9 & 58.2 & 63.4 \\ \bottomrule
\end{tabular}
\caption{Inter-annotator agreement for both the publicly released QueryNER test set and a company-internal agreement trial set. Fleiss' Kappa over all annotators and Cohen's Kappa between pairs of annotators are given at both the entity and token level. }
\label{tab:agreement}
\end{table*}

We computed inter-annotator agreement using Fleiss' Kappa across all three annotators and also Cohen's Kappa between pairs of annotators. 
Agreement measures are given in Table \ref{tab:agreement}.
Agreement was relatively high in the initial internal annotation, but was lower when annotating the publicly released Amazon Shopping Queries Dataset.
Cohen's kappa values for the public dataset were more consistent across annotator pairs than the internal annotation where it appeared annotator 2 tended to conflict more with annotators 1 and 3. 

While the domain of both the internal and public data sets are e-commerce, there could be slight differences in the types of queries which lead to the difference in agreement scores. 
For example, the internal data included more automotive parts and accessories than the Shopping Queries Dataset.
The internal trial annotation also was a smaller amount of data and consisted of only 400 queries with three annotators per query, while the subset of the Amazon Shopping Queries dataset that was annotated for QueryNER included 1,000 queries with three annotators per query for the test set. 
 
\subsection{Dataset}

\begin{figure}[tb]
\large
\centering
\includegraphics[width=\columnwidth]{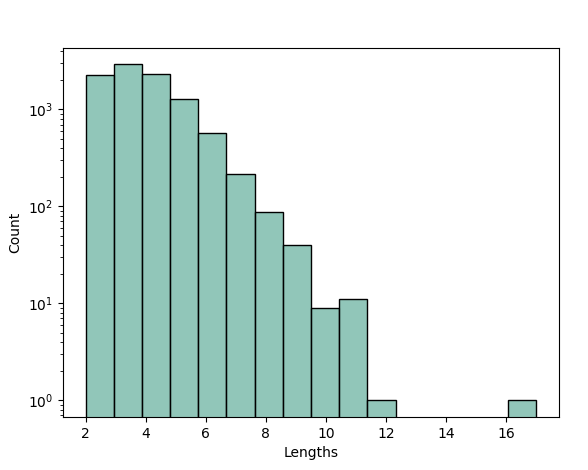}
\caption{Distribution of query lengths with counts on a log scale}
\label{fig:query-lengths}
\end{figure}

\begin{figure}[tb]
\large
\centering
\includegraphics[width=\columnwidth]{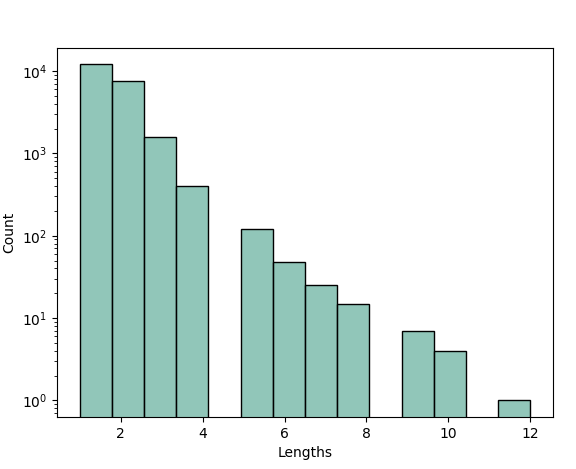}
\caption{Distribution of entity lengths with counts on a log scale}
\label{fig:entity-lengths}
\end{figure}

The final QueryNER dataset contains close to 1,000 queries in the test set and over 7,000 queries in the training set. 
The average of the lengths of all queries is 3.63 tokens, and the distribution of lengths of queries are shown in Figure \ref{fig:query-lengths}. 
The average length of an entity span is 1.60 tokens. 
We also present the count of entities of different lengths in Figure \ref{fig:entity-lengths}. 
Entity lengths are shown on a log scale to show that while the vast majority of entities are one or two tokens long, there are smaller quantities that do have longer lengths.

\begin{table}[tb]
\small
\centering
\begin{tabular}{@{}lrrr@{}}
\toprule
 & Queries & Entities & Tokens \\ \midrule
Train & 7,841 & 17,505 & 28,457 \\
Dev & 871 & 1,930 & 3,124 \\
Test & 933 & 2,317 & 3,610 \\ \bottomrule
\end{tabular}
\caption{Counts of queries, entities, and tokens in each of the QueryNER dataset splits}
\label{tab:stats}
\end{table}

\begin{table}[tb]
\small
\centering
\begin{tabular}{@{}lr@{}}
\toprule
Entity Type & Count \\
\midrule

Core Product Type & 8,310 \\
Modifier & 3,367 \\
Creator & 2,217 \\
Department & 1,652 \\
Product Name & 1,345 \\
Content & 1,301 \\
UoM & 862 \\
Color & 691 \\
Shape & 607 \\
Material & 569 \\
Occasion & 397 \\
Condition & 178 \\
Quantity & 104 \\
Price & 51 \\
Origin & 40 \\
Time & 32 \\
Product Number & 31 \\
\bottomrule
\end{tabular}
\caption{Counts of entity types in QueryNER}
\label{tab:types}
\end{table}

Table \ref{tab:stats} shows the number of queries entities and tokens in each of the train, development, and test splits.
Table \ref{tab:types} shows the balance of entity types in the corpus. 
Unsurprisingly, \texttt{core\_product\_type} is the most frequent type since most queries have a main product. 
The next most frequent types are \texttt{modifier} and \texttt{creator}.

\section{Experiments}
We conducted three sets of experiments.
We set baseline results on the QueryNER dataset. 
We examined the effect of dropping spans identified by QueryNER compared with token dropping as a recovery for null and low recall queries.
Finally we experimented with simple data augmentation techniques to probe the robustness of our models. 
\citet{palen-michel-etal-2021-seqscore} highlighted some sequence labeling  evaluation issues from invalid BIO label sequences. 
We use what they refer to as ``conlleval" repair for all score reporting in this work.
For the following experiments, unless otherwise stated, we use huggingface to implement an encoder with a token classification head with hyper-parameters of a batch size of 16, 20 epochs of training, a learning rate of 5.0e-5, and a warmup ratio of 0.1. 
All experiments are run using 10 different random seeds. 
We report the average precision, recall, and F1 score with standard deviations. 

\subsection{Baseline Tagging Experiments}

\begin{table}[tb]
\begin{tabular}{@{}llll@{}}
\toprule
 & Precision & Recall & F1 \\ \midrule
BERT & 60.94\textsubscript{\textpm0.5} & 60.17\textsubscript{\textpm0.4} & 60.56\textsubscript{\textpm0.4} \\
XLM-R & 60.45\textsubscript{\textpm0.5} & 59.75\textsubscript{\textpm0.5} & 60.10\textsubscript{\textpm0.5} \\
BERT-cont. & 61.78\textsubscript{\textpm0.4} & 60.82\textsubscript{\textpm0.3} & 61.29\textsubscript{\textpm0.3} \\ \bottomrule
\end{tabular}
\caption{Baseline results of BERT, XLM-R, and BERT with continued pre-training on the rest of the ESCI e-commerce dataset.}
\label{tab:baselines}
\end{table}
We establish baseline performance on the QueryNER dataset.
We train a sequence labeling model using BERT\citep{devlin-etal-2019-bert}, XLM-R \citep{conneau-etal-2020-unsupervised}, and also a BERT model with further pre-training using masked language modeling on the rest of the Amazon ESCI queries not already annotated in QueryNER. 
Further pre-training has been shown to increase scores when adapting a model to a specific domain or task \citep{gururangan-etal-2020-dont,lee2020biobert}.

The baseline scores are reported in Table \ref{tab:baselines}.
The baseline scores demonstrate the challenge of segmenting e-commerce queries which have less context and a freer word order than typical English sentences. 
For comparison, performance on the English CoNLL NER task  has F1 scores reported in the nineties (e.g. \citealt{chiu-nichols-2016-named,akbik-etal-2018-contextual}). 
Using a Wilcoxon rank-sum test, the difference between BERT and BERT with continued pretraining on the rest of the Amazon ESCI dataset's queries has a p-value of 0.0008 and is statistically significant despite being a fairly small difference in F1 score.

\subsection{Token vs Entity Dropping}
For this experiment, we attempt applying the segmentation model trained using data annotated with the QueryNER ontology to do query reformulation for recovery on null and low recall queries.  
We consider 5,471 null and low recall queries and segment them with a model trained on an internal version of the QueryNER dataset. 
We then create variants for each query, one by randomly dropping two tokens (\texttt{token-drop-2}) and one which drops a random entity but preserves any entity with the type \texttt{core\_product\_type} (\texttt{entity-drop-keep-core}).

\begin{figure}[tb]
\large
\centering
\includegraphics[width=\columnwidth]{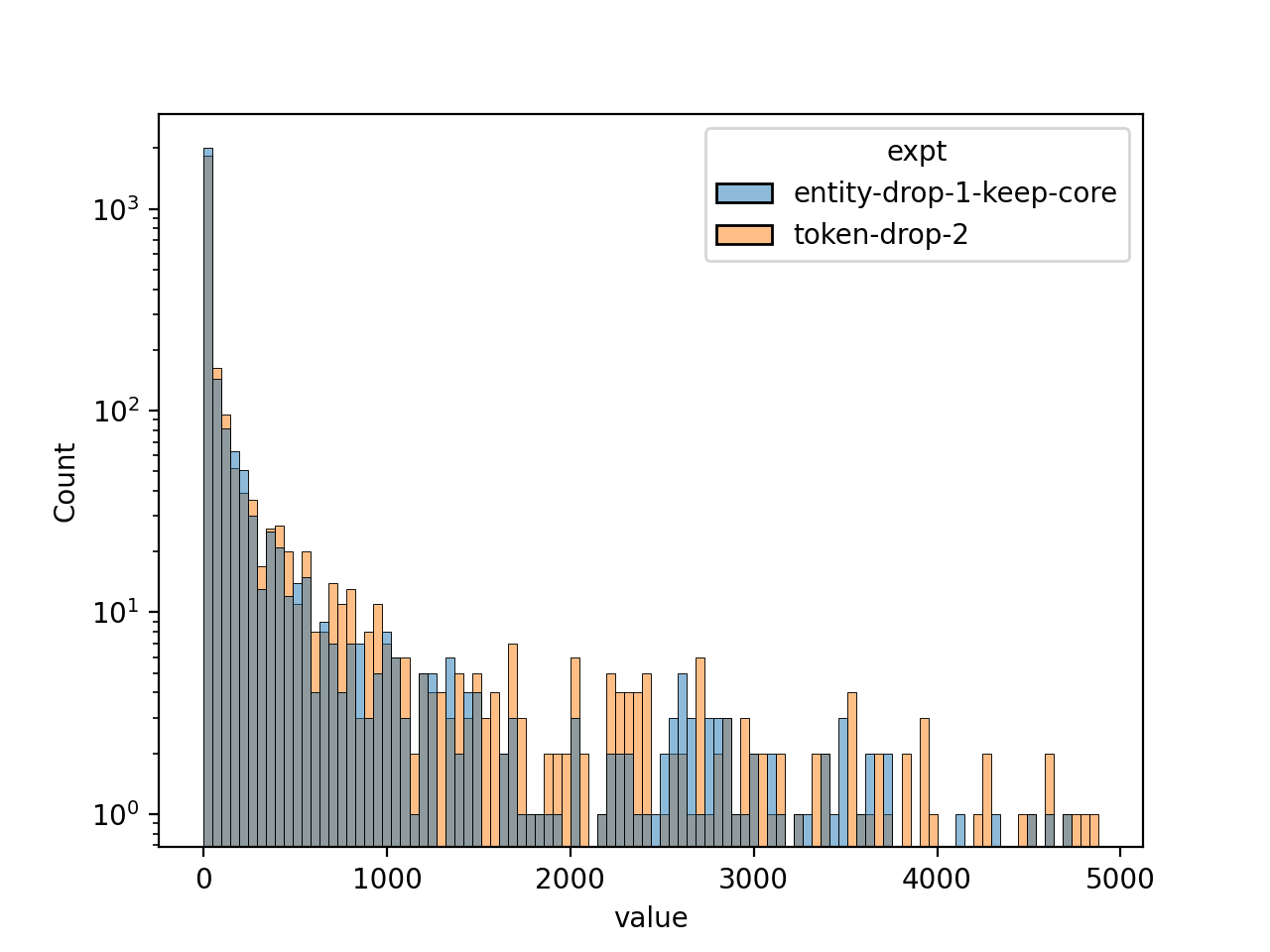}
\caption{Gain in number of items returned from random dropping of tokens vs entities}
\label{fig:recall}
\end{figure}

\begin{figure}[tb]
\large
\centering
\includegraphics[width=\columnwidth]{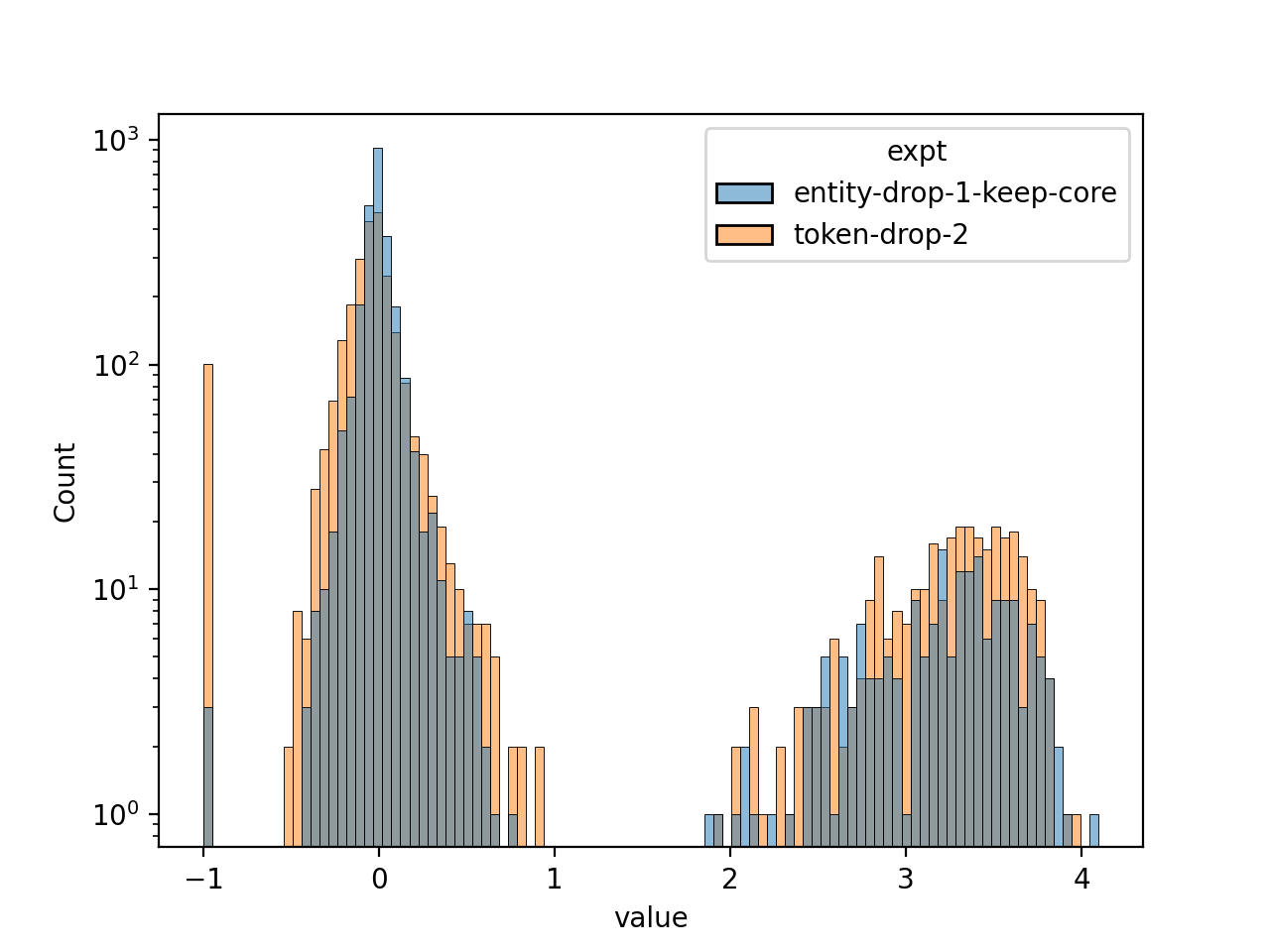}
\caption{Gain in relevance from random dropping of tokens vs entities}
\label{fig:relevance}
\end{figure}

We run the original query and the variants through our internal information retrieval system to get measures of recall and relevance.
Recall is just the number of items returned. 
Relevance is a model-based measure on a scale of 1 to 5, where 5 is most relevant.
Relevance is an aggregate of the top 60 items returned.
We compute the proportional delta between the original  query and the variant from token or entity dropping and show the count of queries binned by their gain in recall (shown in Figure \ref{fig:recall}) and relevance (shown in Figure \ref{fig:relevance}). 

Since the average number of tokens in an entity is 1.6 tokens and so differs from the random dropping of two tokens, these experiments are not directly comparable.
However, overall the mean decrease in relevance over all the queries was lower for dropping an entity and keeping the core product type than for randomly dropping two tokens while maintaining similar gains in recall.

While random dropping of two tokens does appear to have more cases where the gain in recall is substantial, it is possible many of these query rewrites are lower in relevance and relaxing the constraints too much or had very low recall in the original query. 
A proportional gain of thousands of items may suggest that the query has been relaxed too much, though we note that there are few of these cases. 
Note the right most peak in the distribution for the relevance experiment in Figure \ref{fig:relevance} is on a log scale and so there are very few that have such a large relevance increase.
To have that much of a relevance increase the original query would have to have returned very few relevant items.

\subsection{Data Augmentation}
Given the short nature of e-commerce queries and the lack of context available to the model, we hypothesize that the models trained on the QueryNER dataset may not be robust to unseen or noisy data. 
The model may memorize positional information (for example, creators tend to come at the beginning) or may memorize specific tokens as being a certain entity type. 
\citet{lin-etal-2021-rockner} applied transformations to create adversarial examples to make a more challenging test dataset. 
They then showed how training with augmented data could lead to a more robust model. 

\begin{table}[tb]
\centering
\resizebox{\columnwidth}{!}{
\begin{tabular}{@{}ll@{}}
\toprule
Transformation & Example \\ \midrule
Original & airforce 1 women shoes white \\
Shuffled & shoes women white airforce 1 \\
Butterfinger & airvorce 1 women shoes white \\
Numeric & airforce 6 women shoes white \\
Color & airforce 1 women shoes green \\
Mention Replacement & zerogrand boys shoes leopard \\
All Transformations & shofs boys maple zerogrand \\ \bottomrule
\end{tabular}
}
\caption{Examples of data augmentation transformations}
\label{tab:aug-examples}
\end{table}

We similarly apply a series of transformations to the QueryNER test set to create a more challenging test set of queries. 
We assess the best performing model from the baseline experiments, BERT plus continued pre-training on the rest of the Shopping Queries Dataset, on this transformed test set.
We then train individual models using the concatenation of transformed versions of the training data and the original training data. 
We compare how these models trained on augmented data perform on both the challenge transformed test sets and also the original QueryNER dataset.

We apply five transformations to the QueryNER train and test sets. 
Examples of the transformations are shown in Table \ref{tab:aug-examples}.

\textbf{Shuffled:}
Shuffled is simply a random shuffling of the entity spans of the test set. 

\textbf{Butterfingers:}
The butterfingers transformation replaces a small number of characters as if someone has made a typographical mistake.
We use the implementation by \citet{dhole2021nlaugmenter} in their NL-Augmenter package.

\textbf{Color:}
We replace just color spans with other color spans from tagging the rest of the Shopping Queries Dataset and using these entities for replacement.
We limit to only colors with two tokens or less to avoid mislabeled color spans. 

\textbf{Mention replacement:}
We create a set of entity mentions from tagging the rest of the Shopping Queries Dataset. 
We replace entity spans for entities that co-occur with a particular core product type or a particular creator.
For example, we replace the entities around ``shoe" with other entity spans that have co-occurred with ``shoe" in other contexts. 

\textbf{Numeric:}
Replaces number words and digits using the implementation by \citet{dhole2021nlaugmenter}. 

\textbf{All Transformations:}
Applies all transformations in the order of mention replacement, shuffle, butterfinger, numeric, color swap.

\begin{table*}[tb]
\small
\centering
\begin{tabular}{@{}lllllll@{}}
\toprule
Training Data & \multicolumn{3}{c}{Original} & \multicolumn{3}{c}{Transformed + Original Training} \\ 
\cmidrule(r){0-0} \cmidrule(lr){2-4} \cmidrule(lr){5-7} 
\cmidrule(r){0-0} \cmidrule(lr){2-4} \cmidrule(lr){5-7} 
Augmentation & Precision & Recall & F1 & Precision & Recall & F1  \\
\midrule
Butterfinger & 30.86\textsubscript{\textpm0.6} & 33.65\textsubscript{\textpm0.5} & 32.19\textsubscript{\textpm0.5} & 54.59\textsubscript{\textpm0.7} & 54.88\textsubscript{\textpm0.5} & \textbf{54.73\textsubscript{\textpm0.6}} \\
Numeric & 60.80\textsubscript{\textpm0.7} & 58.32\textsubscript{\textpm0.5} & 59.54\textsubscript{\textpm0.6} & 60.45\textsubscript{\textpm0.6} & 59.32\textsubscript{\textpm0.6} & \textbf{59.88\textsubscript{\textpm0.6}} \\
Mention Replacement & 74.83\textsubscript{\textpm1.1} & 75.52\textsubscript{\textpm1.0} & 75.17\textsubscript{\textpm1.1} & 74.88\textsubscript{\textpm0.7} & 77.74\textsubscript{\textpm0.7} & \textbf{76.28\textsubscript{\textpm0.7}} \\
Color & 61.55\textsubscript{\textpm0.6} & 60.62\textsubscript{\textpm0.4} & \textbf{61.08\textsubscript{\textpm0.5}} & 61.37\textsubscript{\textpm0.9} & 60.59\textsubscript{\textpm0.8} & 60.98\textsubscript{\textpm0.8} \\
Shuffled & 59.82\textsubscript{\textpm0.5} & 59.41\textsubscript{\textpm0.5} & 59.62\textsubscript{\textpm0.4} & 64.50\textsubscript{\textpm0.4} & 66.08\textsubscript{\textpm0.5}& \textbf{65.28\textsubscript{\textpm0.4}} \\
All Transformations & 27.70\textsubscript{\textpm0.4} & 30.53\textsubscript{\textpm0.6} & 29.04\textsubscript{\textpm0.4} & 66.25\textsubscript{\textpm0.7} & 69.36\textsubscript{\textpm0.6} & \textbf{67.77\textsubscript{\textpm0.6}} \\ \bottomrule
\end{tabular}
\caption{Data augmentation experiments on transformed test sets}
\label{tab:aug-results}
\end{table*}

\begin{table*}[tb]
\small
\centering
\begin{tabular}{@{}llll@{}}
\toprule
Training Data & \multicolumn{3}{c}{Transformed + Original Training} \\ 
\cmidrule(r){0-0} \cmidrule(lr){2-4}
Augmentation & Precision & Recall & F1  \\
\midrule
No Augmentation & 61.78\textsubscript{\textpm0.4} & 60.82\textsubscript{\textpm0.3} & 61.29\textsubscript{\textpm0.3}\\
Butterfinger & 58.93\textsubscript{\textpm0.5} & 59.78\textsubscript{\textpm0.5} & 59.35\textsubscript{\textpm0.5} \\
Numeric & 60.41\textsubscript{\textpm0.7} & 59.35\textsubscript{\textpm0.7} & 59.88\textsubscript{\textpm0.7} \\
Mention Replacement & 60.04\textsubscript{\textpm0.8} & 59.50\textsubscript{\textpm0.7} & 59.77\textsubscript{\textpm0.7} \\
Color & 60.58\textsubscript{\textpm0.7} & 59.05\textsubscript{\textpm0.6} & 59.80\textsubscript{\textpm0.6} \\
Shuffled & 60.32\textsubscript{\textpm0.5} & 59.97\textsubscript{\textpm0.6} & 60.14\textsubscript{\textpm0.5} \\
All Transformations & 55.65\textsubscript{\textpm0.7} & 54.74\textsubscript{\textpm0.7} & 55.19\textsubscript{\textpm0.7} \\ \bottomrule
\end{tabular}
\caption{Results of models trained on data augmentation data evaluated on the original test set}
\label{tab:aug-results-original-test}
\end{table*}

The results from the augmentation experiments are shown in Table \ref{tab:aug-results}. 
For all experiments we used the BERT model with continued pre-training on the Shopping Queries Dataset as the base model. 
When evaluating the model trained on the original training data on the transformed challenge test sets, we did see a severe drop in performance for butterfinger and all transformations test sets. 
Modest drops in performance occurred for numeric, color, and shuffled test sets.  
The mention replacement test set appeared to become easier rather than more difficult. 
Since the core product and creator from the original dataset were preserved, perhaps these served as sufficient cues to make the dataset less challenging. 

As expected, performance improves on the transformed test sets when training a model on both the transformed training data concatenated with the original training data.
However, even with incorporating transformed data into the training, the butterfinger transformation test set still appears to be fairly challenging. 

The performance of all the models trained on the combination of original and transformed datasets all perform slightly worse than the baseline models trained only on the original training data.
It is notable, however, that despite this roughly two points of F1 degredation in performance on the original test set, there is notable gain in the challenge transformed test sets for the butterfinger transformation, shuffled transformation and all transformations applied at once. 
These experiments demonstrate that if the goal is to produce a robust segmentation model that performs well even under cases with spelling mistakes and free word order, training on a combination of augmented data can help, but at the cost of a slight drop in performance on clean less noisy data.

\section{Discussion}

The results of our annotation efforts show that the task presented by QueryNER is challenging.
We were able to achieve relatively high agreement despite very little contextual information for the annotators.
F1 scores on baseline experiments provide room for further improvement, and we showed through augmentation experiments that simple data augmentation strategies can make the models trained on QueryNER more robust to noise.

\subsection{Limitations}
There were a number of challenges and limitations in defining the QueryNER ontology and creating the dataset. 
For example, determining when to consider a token as a core product type with a modifier \texttt{[hiking][boots]} compared with a single chunk \texttt{[hiking boots]} as a core product type.   
We favored using the longer span of \texttt{[hiking boots]}, but there are cases of greater ambiguity with longer noun-noun compounds such as ``dog cone collar''.
It can be difficult for annotators to decide whether the correct chunking should be \texttt{[dog cone] [collar]}, \texttt{[dog][cone collar]}, or \texttt{[dog cone collar]}. 
Many of the annotator disagreements come from slight span mismatches despite guidelines urging annotators to prefer longer more complete spans. 
There are also a number of ambiguous tokens that can be difficult to differentiate between types. 
Consider ``mazda'' which can be both a \texttt{creator} as the make of a car, but also appears as a \texttt{product name} in the phrase ``mazda 3''.

Another limitation is that the dataset is only in English, though we did a small internal experiment showing the promise of multilingual transfer to Spanish with similar performance as the English data. 

The dataset was created with a specific set of business use cases in mind. 
The underlying shopping queries are from the Shopping Queries Dataset and therefore is only representative of the types of queries included within it.
When applying the ontology we created based on internal data to the public Amazon Shopping Queries dataset, there were differences in the distribution of entity types from the public data. 
The internal data had more years and dates and more frequently discussed the condition of an item, so the entity types \texttt{time} and \texttt{condition} were more prominent in the internally annotated data.
While the ontology appeared to cover the public shopping queries dataset and was tested internally on multiple shopping categories, other subdivisions of entity types may be useful for other use cases or specific categories.
We would expect the models trained on the QueryNER dataset to be somewhat limited by the types of queries available within this dataset.

\subsection{Future Work}
There are a number of promising directions for future work. 
We have done some internal annotation of product titles showing broad applicability of the ontology, a potential direction for future work could be to conduct more annotation of product titles on a public dataset. 
Another potential direction of future work could be to add relations between chunks of queries or product titles, clustering the chunks, or linking them to a knowledge graph.
The Shopping Queries Dataset includes Spanish and Japanese as well. 
We experimented internally with transfer learning on a small set of 100 queries, but another promising direction would be to annotate queries from other languages.

\section{Conclusion}

We defined a type ontology and annotation guidelines that are broadly applicable to e-commerce segmentation and  
released QueryNER, a new manually annotated dataset and open benchmark for query segmentation. 
Our baseline models showed that the task of e-commerce query segmentation is challenging due to lack of context from short strings of text.
We showed one promising direction for using QueryNER as part of a null and low query recovery strategy by dropping spans rather than individual tokens.  
Experiments with data augmentation showed how baseline models are not robust to transformations and noise, especially to permutations at the character level within a word. 
We showed that using artificially augmented training data can help the model to be more robust to this type of noise, but at a slight cost of performance when measuring on the original test set.

\section{Ethics and Broader Impact}
We have attempted to design an ontology that is broadly applicable in e-commerce queries and product titles and have tested using it with a range of different product categories. 
The annotation is inevitably the product of the biases and opinions of the designers of the ontology and the annotators. 
We have made efforts to report agreement measures and will release the original annotations of each annotator before adjudication for transparency. 

The annotation effort was salary and contract work and there was no specific hourly wage.  But annotators were paid a minimum living wage. Crowd sourcing was not used for this work. 

We believe the impact of QueryNER will be positive since there are few if any publicly available chunking or NER datasets in the e-commerce domain.

\section{Acknowledgments}
This work was supported by the grant ``Improving Relevance and Recovery by Extracting Latent Query Structure'' by eBay to Brandeis University.

\section{Bibliographical References}
\label{sec:reference}

\bibliographystyle{lrec-coling2024-natbib}
\bibliography{custom}

\section{Language Resource References}
\label{lr:ref}
\bibliographystylelanguageresource{lrec-coling2024-natbib}
\bibliographylanguageresource{custom_language_resource}

\appendix 

\section{QueryNER Annotation Guidelines}
\label{sec:guidelines}

\subsection{Overview}
The goal of this task is to divide a user’s query into meaningful chunks and assign a broad type to each span. 

As an example:

\texttt{\textcolor{TealBlue}{[High - end]}}  \texttt{\textcolor{blue}{[speaker cover]}}   \texttt{for}   \texttt{\textcolor{CadetBlue}{[B \& W]}   \textcolor{RedOrange}{[805d]}  \textcolor{RedViolet}{[1 pair]}  \textcolor{Fuchsia}{[made of velvet]}   \textcolor{OliveGreen}{[suede]}}  \texttt{\textcolor{Brown}{[made to order]}}

``Speaker cover'', ``made of velvet'', ``made to order'', ``1 pair'' and ``high-end'' are all natural chunks of the query. The goal of this task is to better understand a query by being able to better break it up into meaningful pieces and assigning meaningful types to the spans. For example ``made of velvet'' clearly refers to the material while ``made to order'' is a more general description of the product. 

\subsection{Motivation}
The goal of these annotation guidelines is to provide a type ontology for sequence labeling of e-commerce queries and product titles into meaningful chunks. While this task is similar to extraction of e-commerce aspects, there are some notable differences.

Unlike the e-commerce task of aspect extraction, which tends to focus on fine-grained types that are often specific to particular categories of items, the tag types of QueryNER aim to be broadly applicable to queries for any product category. The goal is not to extract aspects, but rather to segment the user's query into meaningful chunks. 

The type ontology is also meant to be small and general purpose enough that annotators do not necessarily need to become domain experts in the products involved in the annotation process. 

\subsection{Steps}
\begin{enumerate}
\item First familiarize yourself with the tag types and procedures in this document. 
\item Read the full string of text first.

\item Try to identify the main entity that is being described. For example, ``sneakers''. This will often be core\_product\_type but may be other types if there is no generic product mentioned. If it is unclear what the main entity being sold is, it is best to do a search to try to find the item or similar items. This can also help identify unfamiliar brand names or item specific terminology.

\item Mark the spans that are clearest or easiest to identify first. It is not required to go from left to right. This can help narrow down the more difficult decisions.
\item However, you may need to reconsider these decisions after doing a web search as some things that seem like obvious design descriptions or demographics may actually be part of an unfamiliar brand name. 
\end{enumerate}

\subsection{Rules \& Tips}
\begin{itemize}
\item Assign a tag to every word in the query.

\item Pay attention to context. For example “gold” would be tagged as Material in a fine jewelry item, but as Color for a pair of sandals.
\end{itemize}

\textbf{No Tag}

\begin{itemize}

\item Use “No Tag” for words or punctuation that is clearly not part of any span. 
\item Using “No Tag” for this task should be very rare and should be avoided as much as possible.

\item Tag special characters when they are part of a chunk:
The ``\&'' in "Abercrombie \& Fitch" should be tagged as "Brand Name" because it is part of the trademarked brand name. 

\item The “\&” in ``blue \& green'' should get “No Tag” because it serves as punctuation only since “blue” and “green” can serve as their own chunks.

\item Do not tag prepositions when they are just joining two core\_products. For example, [Earbuds] with [case]. You should include the prepositions when they are a core part of the span of a modifier: [Earbuds with bluetooth] [with mic]. Do not tag prepositions with “No tag” when they are part of a chunk of the query.
\end{itemize}

\textbf{Obscure tag}
Use the ``Obscure'' tag for words that you cannot decipher.
Words and titles not in the Native Language (except for EN) should be tagged as “Obscure” unless:
\begin{itemize}
\item The word is commonly used in the native language. For instance: "attaché" and "Art Nouveau".

\item The word is part of a brand name or a product name. For instance, the French brand name "Petit Bateau".

\item If the majority of words are not in the native language (except for EN), all words should be marked as “Obscure”.

\item Tag misspellings and abbreviations whenever possible. If you cannot understand the meaning of the word or special character, use “Obscure”.

\item If a word has a gender disagreement or the word order is not fluent, tag it normally. 
For example, “iluminado retículo” instead of “retículo iluminado” or “objetivo” instead of “objetiva”.
\end{itemize}

\textbf{Spans}
Tag the full span of tokens for each chunk
Always attempt to mark the most complete span possible. 
For each meaningful chunk of a query, the entire span should be marked. For example, for the text “hiking boots made in Italy size 12”, “hiking boots” should be marked core\_product\_type rather than two separate spans.  “made in Italy” should be marked ``Origin''. Note the entire span is marked and not just “Italy”. “Size 12” is also marked in its entirety as UoM rather than just marking the span “12”. 

\textbf{How to tell whether to divide a chunk further?}
If removing a portion of a chunk fundamentally changes the meaning, it should be kept as a single chunk. For example, [wedding dress] vs [dress] is a different product. Another test is to reword to separate the two spans. [dress] [to wear at a wedding] is a different product than an actual wedding dress since it could be a bridesmaid’s dress or a guest’s dress. Another example, [ausdom] [headphone pads] would mark “headphone pads” as a span  since just “pads” could include brake pads, pads for furniture, or other types of pads. 24v car battery charger

\textbf{Separate distinct chunks}
It is possible for tags of the same type to appear consecutively. These should be divided into separate chunks. For example, “looney tunes bugs bunny daffy duck coffee mug” should be tagged as 
``\textcolor{red}{looney tunes} \textcolor{ForestGreen}{bugs bunny} \textcolor{blue}{daffy duck} \textcolor{purple}{coffee mug}''. 
Where ``\textcolor{red}{looney tunes}'', ``\textcolor{ForestGreen}{bugs bunny}'', and ``\textcolor{blue}{daffy duck}'' are separate chunks tagged as ``content'' and ``\textcolor{purple}{coffee mug}'' is tagged as core\_product\_type. 

\subsection{Comparisons of Types}

\subsubsection{Core\_product\_type vs Modifier}
Examples like “hiking boots”, “battery tray”, or “wedding dress” are tagged as a single core\_product\_type since the multi-word span describes a unique product. For example, with “wedding dress” vs ”dress” there is a greater change in meaning, where  for “comfortable boots”, comfortable should just be marked as a modifier as it is not part of the core meaning of the product. 

Here [two way] is a modifier while [car speaker] is a core\_product\_type.
[Two way] [car speaker]

\subsubsection{Creator vs Product\_name}

Product\_name is not to be used for brand names or names of companies, which are marked creator. 
Kia [creator] sorento [product\_name]

\subsubsection{Product\_name vs Product\_number}

Product names and product numbers can occur in the same query. Product numbers differ in that they are more of an identifying number and not a known product line or specific name, while product names may contain numbers, but typically have other non-number components like “Campus 80s” in the example below or “f20”. FW7619, however, is an alphanumeric identifier for the product’s style. 
Adidas[creator] Campus 80s[product\_name] By Alex Nash[creator] FW7619[product\_number] 

Even product names, models or lines of products that contain numbers are marked product\_name and not product\_number. product\_number is used for numbers that are not names but identifiers or codes used by either eBay or another retailer. “frame” is marked as core\_product\_type because it is the main item being sold and is a generic mention and not a specific name of a facemask frame. 
Resmed [creator] f20 [product\_name] frame [core\_product\_type]

The number for a product. This can be an eBay product number or a company’s product number. Note that it cannot be a model number that is used as the name of a product, as in Ford F150, F150 is the product’s name.

\subsubsection{Condition vs Modifier}

Condition refers to the quality, newness, or availability of the product. More general descriptions of how the product will be made are left to the modifier label. For example, High-end[condition] 
vs made to order[modifier].

High-end[condition] speaker cover for B\&W 805d 1 pair made of velvet suede made to order[modifier]

If you encounter one you aren’t familiar with, it is often possible to find them explained by doing a web search. 
eBay maintains a list of acronyms: 
\url{https://www.pages.ebay.com/pr/es-co/help/account/acronyms.html}
However you may find some that aren’t included in that list. It is usually possible to find the meaning with a web search. 
This website also has a list of e-commerce acronyms:
\url{https://resellingrevealed.com/ebay-abbreviations-acronyms/}

\subsubsection{Department vs UoM}

For departments  next to sizes, still mark them separate even though the span may be indicating a set of sizes based on gender or age group.  So “men’s size 12” should still be marked department and UoM.

\subsubsection{Occasion vs Core\_product\_type}
`hiking boots' is its own product and `hiking` in this case should NOT be marked as a purpose. Only mark purpose when the span is not a typical phrase to describe a product, for example, “plates for [wedding]”

\subsubsection{Time vs UoM}
Time that is used as a measurement of how long something lasts should be marked as a unit of measurement (UoM), but should be marked as time when not used as a measurement. Time periods, such as 1912-1915 should still be marked as time despite being a duration. 

[4 hours] laptop battery
“4 hours” is marked as UoM since it is a measurement of the battery life.

Battery from [1910]
“1910” is marked as time since it is not a measurement of the battery.

\subsubsection{Time vs Demographic}
In this case, teens refers to the time “1910s” rather than a demographic. 
Teens[time] 1920s[time] 

\subsubsection{Content vs Creator}
Items that are signed are considered content as well as the name of the person signing if included. Note that the name from a signature differs from the name of the creator. In cases where the signer is also the creator, mark as creator. 

\subsubsection{Quantity vs Condition}
Note just because a number is referenced does not mean it is a quantity. In the example below, 4 unopened is marked as Condition since it refers to the condition of a portion of the items.
[Lot of 6] plastic canvas kits Christmas Needlecraft Shop [4 unopened]

\subsubsection{Origin vs UoM}
Be careful that this is not part of the size. Clothing often specifies the country for the size, for example “USA size 12”. Since this whole span is a size it should be marked as a unit of measurement [UoM]. 

\subsubsection{Origin vs Content}
Locations may also show up as content on items like clothing or posters. In these cases, they should be marked content. 

\subsubsection{Origin vs Creator}

Origin locations may also appear in product names or names of companies. The whole span of text “Nintendo of America” would be marked as creator rather than origin. “Milwaukee” is a brand name of tools and should be marked creator. 

\onecolumn
\begin{center}
\begin{longtable}{p{0.18\linewidth} p{0.35\linewidth}p{.35\linewidth}}

\toprule
Tag &
Definition &
Examples \\
\midrule
core\_product\_type & The main thing being sold. Generic ways of describing a product. These are not official product names but common objects. & teapot, tennis shoes, figurine, lounge pants, dish soap, comic book \\
\midrule
product\_name & The specific name of a product or model name & F150, air jordan 7, sorento \\
\midrule
product\_number & The number for a product. Can be an eBay product number or a company's product number. Editions of an item that are numbered can also be marked as product number as well as trading card or comic numbers. & BQ4422-001, 7101, airforce 1 cw2290-111 size 12. In this example, cw2290-111 is used as a style code. This particular item also has an eBay product id of 6045644602, which would also be marked as product\_number if it was included in the text of the query. \\
\midrule
modifier & Modifier is used for spans that clarify the type of product. This can describe certain features a project has like “2 in 1” or “high performance”. Modifier can also be used for constraining the type of a product. Modifier can also be used as a catch all for “type” of a product that doesn’t fit in other predefined categories & 4 in 1 system, performance, essential, 1 spray, easy, garage.

For “car battery charger” “car” modifies the “battery charger” which would be tagged core\_product\_type.

For example, “sport utility” when describing a vehicle as in “kia sportage sport utility”. 
However, if the text is selling a “2011 sport utility vehicle” or “grey SUV”, “SUV” and “sport utility vehicle” would just be marked as core\_product\_type because they refer to the item itself rather than specifying the type of item being sold. 

\makecell{
\\Vintage wooden sculpture Nite owl \\
design statue \textcolor{blue}{[signed]}\\ \\
\textcolor{blue}{[Hi temp]} masking plugs} \\
\midrule
condition & The condition of the product. This describes whether the product is new or old and can go into more detail about things such as whether a product includes its original tags.
There are a number of acronyms commonly used by sellers (and less frequently by buyers).
Also use this tag for descriptions of what the product doesn’t come with or how it was packaged. For example, “no key”, “loose”, “no box”, “in original box”. & new, used, mint condition, vintage, near mint, no box \makecell{
\\NOS = New Old stock\\
OEM = Original Equipment \\Manufacturer\\
EUC = excellent used condition\\
nm = near mint\\
NIB = New in box Condition\\
Vtg = Vintage\\ \\}\\
\midrule
UoM (Unit of Measurement) & Any way of measuring size or other unit of measurement. This can include everything from clothing sizes, to lengths and widths, car engine sizes, battery capacity, amount of memory in a computer, lens sizes for cameras. This includes time expressions that are units of measurement such as 30 minutes or 4 hours of battery life. & \makecell{16oz, L, Medium, 5 inches, 30 mins\\ \\
\textcolor{blue}{[4 hours]} of battery life\\ Tesla model x \textcolor{blue}{{[}12 volt{]}} battery \\ 2020 Shirt FR0820 Men’s \textcolor{blue}{[size S]} new}\\
\midrule
department & Category of the population the item was made for. & \makecell{Mens, womens, kids, jr., wmns\\ Elf clown elk hat for \textcolor{blue}{{[}adults{]} {[}kids{]}} gift}\\
\midrule
material & The material or physical entity that makes up the item. & \makecell{denim, canvas, plastic, metal, cotton, \\felt\\\\ motorcycle jacket black \textcolor{blue}{[leather]}\\ size xl \\ \\ 
VANS hi top black \textcolor{blue}{[canvas]} \\skateboarding \\ shoes\\\\ EASECASE Custom made \\\textcolor{blue}{[genuine leather]} case} \\
\midrule
time & An expression of the date or time associated with the product that is not a unit of measurement. For example, 30 mins or 4 hours for battery life should be labeled UoM. There are time abbreviations to be aware of. 
MCM = Mid Century Modern & \makecell{1999, 2012-13, ‘67, autumn, \\ \textcolor{blue}{[Mid century]} hat ladies }\\
\midrule
content & Names of characters, titles of tv or movies, sayings or phrases that appear on or within the product itself. 
Many mugs, t-shirts, figurines, or comic books have some form of content or characters associated with them. Often multiple consecutive chunks are found for content.
Meaningful chunks or phrases should still be split. & \makecell{Star wars, knights of the old republic\\ big trouble in little china, big bang \\theory, linus, lucy, far east tour\\  \\ \textcolor{blue}{[Boston Marathon 2020]}\\ \textcolor{blue}{[Rise Up N Run]} Shirt } \\
\midrule
creator & The company or person who creates or produces the product. It could also be the designer name associated with the product or brand name. & Ford, Disney, Jim Shore, Honda, Hot wheels, dc comics, polaroid \\
\midrule
color & Description of the color, pattern, appearance related to the surface appearance or 2-dimensional design of the product. 
Color can include colors described by words  that can also be flavors or foods in other contexts. & \makecell{unc blue, light gray, wolf gray, floral\\ 
\\SPENDOR AUdio SP2 Loudspeaker \\\textcolor{blue}{[Cherry]} 
\\\\Color also includes patterns:\\ The emily \& meritt \textcolor{blue}{[pirate stripe]} \\sheet set} \\
\midrule
shape & Description of the shape, form, or positioning or 3-dimensional design of 
the product.
Includes design descriptions for things like clothing, accessories, or automotive that refer to 3-dimensional descriptions of the item. & fit, slim, low, long sleeve, flat, rear, front, rectangular, orb, cylindrical \\
\midrule
quantity & The number of the product being sold. Includes for example “lot of 4”.

Box break: Multiple people split the cost of a box and share the contents & cds, lot of 4, multi-lot, package of 6, 3-box break, Snap-On Tools Hammer \textcolor{blue}{[a set of 4]} great condition \\
\midrule
occasion & The purpose or intended use of an item. Typically an event, holiday, season, or occasion. `hiking boots` is its own product and `hiking` in this case should NOT be marked as a purpose. Only mark occasion  when the span is not a typical phrase to describe a product, for example, “plates for wedding” would be an occasion, but “wedding dress” would just be a core\_product\_type. & sport, athletic, wedding, winter, halloween, bridal, birthday, \textcolor{blue}{[Christmas]} Santa Claus figurine nutmeg shaker, Japanese \textcolor{blue}{[kitchen]} chefs knife moritaka hamono aogami super kurochi 240mm, Elf clown elk hat for adults kids \textcolor{blue}{[gift]} \\
\midrule
origin & The origin of a product. This is likely the location it comes from but could also be a specific event where the item was created such as a convention. States or provinces can also be an origin tag. For example: Amish style kitchen table \textcolor{blue}{[made in Ohio]}  & Made in the USA, USA, Japan, Germany, Comic-con \\
\midrule
price & The price of a product.  Also includes words expressing relative price like “expensive”, “cheap”, or “good deal”. & \$12.99, cheap, affordable, expensive \\
\midrule
no\_tag & Used for punctuation or words that are not part of a chunk. For this task this tag should be used very rarely since the goal is to separate each query into chunks. Each word is expected to be part of some chunk. & ‘,  “,  \&, – \\
\midrule
obscure & Indecipherable text or text in a language outside the target. Please make the best effort when possible to identify spans that may be product numbers or model names or numbers & Asdfjalksjdf, other languages \\
\bottomrule
\caption{Entity tags with their descriptions and examples.}
\label{tab:ontology}
\end{longtable}
\end{center}
\twocolumn

\end{document}